\RequirePackage{fix-cm}
\documentclass[a4paper,11pt,twocolumn,twoside]{article}
\usepackage[dvipdfm]{graphicx}
\usepackage{bmpsize}
\usepackage{sepln_en}
\usepackage{fullname}
\usepackage[utf8]{inputenc}
\usepackage[english]{babel} % La configuración en ingés ya tiene todas las opciones extra del castellano.

\input epsf

\usepackage[dvipsnames,svgnames,table]{xcolor}
\usepackage{color}

\usepackage{caption}
\usepackage{subcaption}
\usepackage{comment}

\usepackage{tabularx, ragged2e, array, arydshln,adjustbox, booktabs, tablefootnote, multirow}
\title{Medical Argument Mining: Exploitation of Scarce Data Using NLI Systems}

\author {\textbf{Maitane Urruela,$^1$} \textbf{Sergio Martín$^2$} \textbf{Iker De la Iglesia$^1$} \textbf{Ander Barrena$^1$} \\
 $^1$HiTZ Center- Ixa, University of the Basque Country UPV/EHU\\
 $^2$University of the Basque Country UPV/EHU\\
 $^1$\{maitane.urruela, iker.delaiglesia, ander.barrena\}@ehu.eus \\
$^2$smartin128@ikasle.ehu.eus 
}

\seplntranstitle{Minería de argumentos médicos: explotación de datos escasos mediante sistemas NLI}

\seplnclave{Medicina Basada en Pruebas, Minería de Argumentos, Procesamiento del Lenguaje Natural.}

\seplnresumen{Este trabajo presenta un proceso de Minería de Argumentos que extrae entidades argumentativas de textos clínicos e identifica sus relaciones argumentativas utilizando técnicas de clasificación de tokens e Inferencia del Lenguaje Natural. En comparación con otros métodos como la clasificación de texto, esta metodología muestra mejor rendimiento en contextos con escasez de datos. Al evaluar la eficacia de estos métodos en la identificación de estructuras argumentativas que apoyan o refutan posibles diagnósticos, esta investigación sienta las bases para futuras herramientas que puedan proporcionar justificaciones basadas en pruebas para conclusiones clínicas generadas automáticamente.}

\seplnkey{Evidence-Based Medicine, Argument Mining, Natural Language Processing.}

\seplnabstract{This work presents an Argument Mining process that extracts argumentative entities from clinical texts and identifies their relationships using token classification and Natural Language Inference techniques. Compared to straightforward methods like text classification, this methodology demonstrates superior performance in data-scarce settings. By assessing the effectiveness of these methods in identifying argumentative structures that support or refute possible diagnoses, this research lays the groundwork for future tools that can provide evidence-based justifications for machine-generated clinical conclusions.}

\firstpageno{1}

\begin{document}

% la siguiente instrucción sólo se debe usar si el abstract sobrescribe el texto
% la longitud variará según se necesite

%\setlength\titlebox{20cm}

\label{firstpage} \maketitle

\section{Introduction}

In recent years, there has been a growing interest in developing intelligent systems to assist healthcare professionals, particularly in the field of Evidence-Based Medicine (EBM). EBM systems aim to extract pertinent information from unstructured clinical documents and transform it into a structured, machine-readable format, enabling automated analysis. Argument Mining (AM), aligning with EBM, examines the evidence and reasoning clinicians use in clinical cases. This process involves identifying argumentative structures within texts---specifically, finding \textit{claims} (a point to be proved) and \textit{premises} (evidence that supports or refutes a \textit{claim}), and establishing \textit{support} or \textit{attack} relations between them. In the clinical context, this process enables the extraction of logical relationships that justify clinical decision-making \cite{stylianou2021transformed}.

\begin{figure*}[ht]
    \centering
    \includegraphics[width=0.93\linewidth]{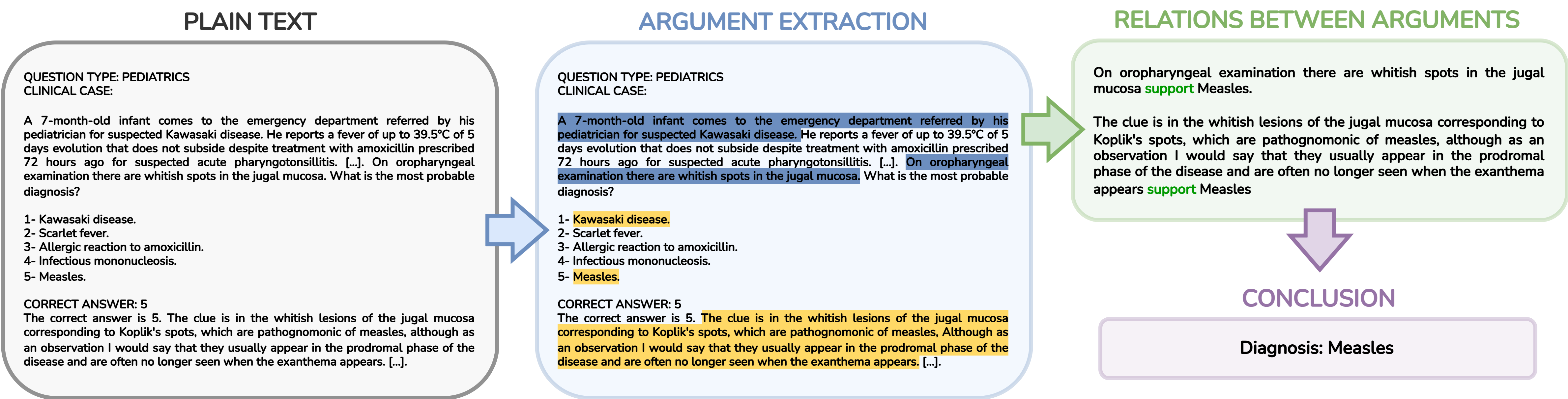}
    \caption{AM system where arguments---\textit{premises} highlighted in dark blue and \textit{claims} in yellow---are extracted (light blue) and their relations are identified (green). Possible applications like final conclusions (purple) are not given in this work.}
    \label{fig:process}
\end{figure*}

However, a significant challenge in advancing these systems is the limited availability of medical corpora, primarily due to patient privacy concerns. The scarcity of data makes it difficult to train and test robust systems, necessitating strategies that maximize the value of the limited data available. In the case of Argument Mining, which involves argument recognition and relation extraction (Figure \ref{fig:process}), the latter is particularly challenging and achieving high performance in it with few examples is challenging even for human annotators.

To address this, cross-task transfer learning offers a promising solution. By fine-tuning pre-trained models, such as those trained on Natural Language Inference (NLI) tasks, with medical domain-specific data, we can leverage its existing argumentative knowledge. This approach allows the model to be adapted for medical argumentation tasks with relatively few examples, which is far more efficient than training models from scratch using end-to-end text classification methods.

\begin{figure*}[ht]
    \centering
    \includegraphics[width=0.8\linewidth]{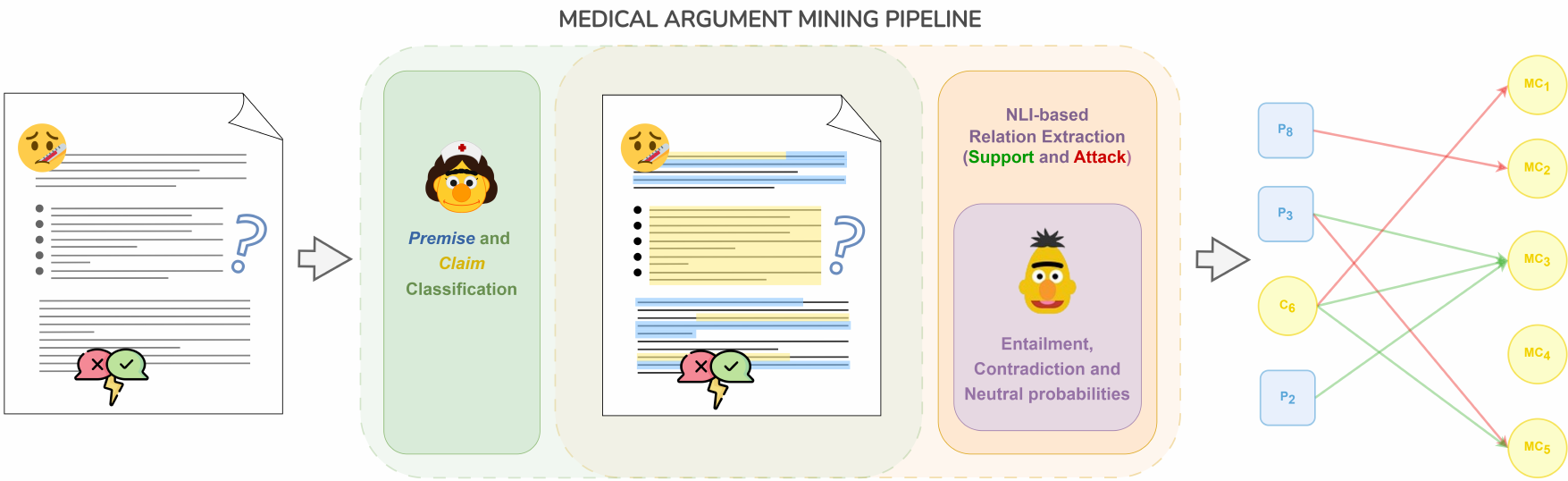}
    \caption{Diagram of the proposed medical AM system. In the argumentative-entity recognition step (green), \textit{premises} and \textit{claims} within the text---containing a clinical case and question, possible answers and a final explanation---are identified. These entities are then processed by the Ask2Transformers library (orange), which leverages entailment probabilities from a fine-tuned NLI model (purple) to infer the underlying relations between argumentative entities.}
    \label{fig:medical_pipeline}
\end{figure*}

This work presents a methodology for Argument Mining that extracts argumentative entities from clinical texts and identifies their relationships using token classification and NLI-based techniques (Figure \ref{fig:medical_pipeline}). This methodology aims to enhance performance in data-scarce environments by tapping into pre-existing knowledge, thereby facilitating more effective extraction of argumentative structures in clinical texts. By evaluating the performance of these techniques in identifying arguments that support or challenge potential diagnoses, this research lays the foundation for future tools that offer evidence-based justifications for machine-generated clinical conclusions. Such tools could provide valuable insights to clinicians, not only by alleviating their workload and improving patient response times but also by accelerating the diagnostic process, ultimately facilitating early health risk detection and prevention.

To meet these objectives, this study addresses the following research questions:
\paragraph{RQ1} Does token classification perform well in argument extraction tasks?
\paragraph{RQ2} Is NLI useful as an intermediate task for argument-relation classification in scarce-data technical domains like medicine? 

\paragraph{RQ3} Does an approach that incorporates NLI-based knowledge outperform a basic text classification strategy for argument-relation classification?

\section{Related work}

Argument mining (AM) involves analyzing discourse at the pragmatics level using argumentation theory to model and automatically analyze data \cite{habernal2017argumentation}. It consists of two main stages:
\paragraph{Argument Extraction} This stage detects arguments within the given text, identifying the boundaries of each argumentation entity in the process. Early works such as \cite{teufel2009towards}, pioneered by analyzing scientific articles, introducing the concept of argumentative zoning. Subsequent research included methods to automatically detect context-dependent claims \cite{rinott2015show,levy2014context} and supporting evidence \cite{bar2017stance} in Wikipedia articles using methods like Logistic Regression methods. Naive Bayes classifiers were also utilized in political texts \cite{duthie2016mining}. Recent studies also use LLMs to generate arguments from texts, for example by extracting question–response pairs \cite{piriyatamwong2024unified}.

\paragraph{Relation Prediction} This stage predicts the relationships between identified arguments, which helps structure the discourse. These relationships, such as \textit{attack} or \textit{support}, are identified using various approaches, including standard SVMs \cite{daxenberger2017essence} and Textual Entailment \cite{cabrio2013natural}. These methods are used to build argument graphs and predict the connections between argumentative entities in structured argumentation.

End-to-end AM systems, like those proposed by \cite{nguyen2018argument} and \cite{eger2017neural}, have been developed to directly parse argumentative structures from free text or persuasive essays. In the medical domain, \cite{enhancingmayer2021} proposed an end-to-end pipeline to address argumentative outcome analysis on medical trials, using MEDLINE database and deep bidirectional transformers combined with different neural architectures (LSTM, GRU and CRF) for an AM process and performing an outcome analysis module based on Population, Intervention, Comparator and Outcome entities (PICO). They obtain a 0.87 Macro F1-score for argument extraction and 0.62 Micro F1-score for \textit{attack} and \textit{support} relations prediction. Similarly, \cite{stylianou2021transformed} introduces an end-to-end SciBERT-based system, which uses PICO entities to enhance the performance of the AM system trained on the dataset of \cite{mayer2020transformer} referred as AMCT. This system achieves a 0.92 Micro F1-score for entity extraction and 0.6 Micro F1-score for \textit{attack} and \textit{support} relations prediction. Additionally, \cite{elenayserenamarro2023automatic} used the Human Phenotype Ontology (HPO) and sentence embeddings to assess the relevance of reasons in medical explanations.

In this work, we propose a two-step system which first leverages a token classification method for medical argumentative entity recognition, followed by a two-step Natural Language Inference (NLI) approach for relation extraction. This second step uses a Large Language Model (LLM), fine-tuned on medical NLI tasks, to reformulate the argument relation classification task (\textit{attack}, \textit{support}, or \textit{no-relation}) as a zero-shot task using Ask2Transformers system \cite{sainz2021ask2transformers}. In \cite{sainz2021ask2transformers} the system is utilized to exploit the domain knowledge of the pre-trained Language Models to enrich the WordNet \cite{fellbaum1998wordnet} synsets and glosses with domain labels. In \cite{sainz-etal-2022-textual} they also use this library to cast Event Argument Extraction (EAE) as an entailment task.

\section{Materials and Methods}
The objective of this project is to develop a two-step tool for Argument Mining (AM) in clinical texts, focusing on identifying argumentative entities and their relationships (Figure \ref{fig:medical_pipeline}). The two distinct steps are: (1) argumentative-entity recognition, where a token classification approach is employed to detect \textit{premises} and \textit{claims} (argumentative entities) within clinical texts; and (2) entity-relation extraction, in which a Natural Language Inference (NLI) model is fine-tuned within the medical domain as an intermediate step for text classification, utilizing the Ask2Transformers library. Both components are evaluated independently to avoid error propagation---in other words, the relation extraction stage uses gold-standard entity annotations rather than relying on outputs from the entity recognition step. 

This section provides a comprehensive overview of the materials and methodologies used to accomplish these tasks.

\subsection{Materials}
\subsubsection{Dataset} \label{sec:dataset}

\begin{figure}[ht]
    \centering
    \includegraphics[width=\linewidth]{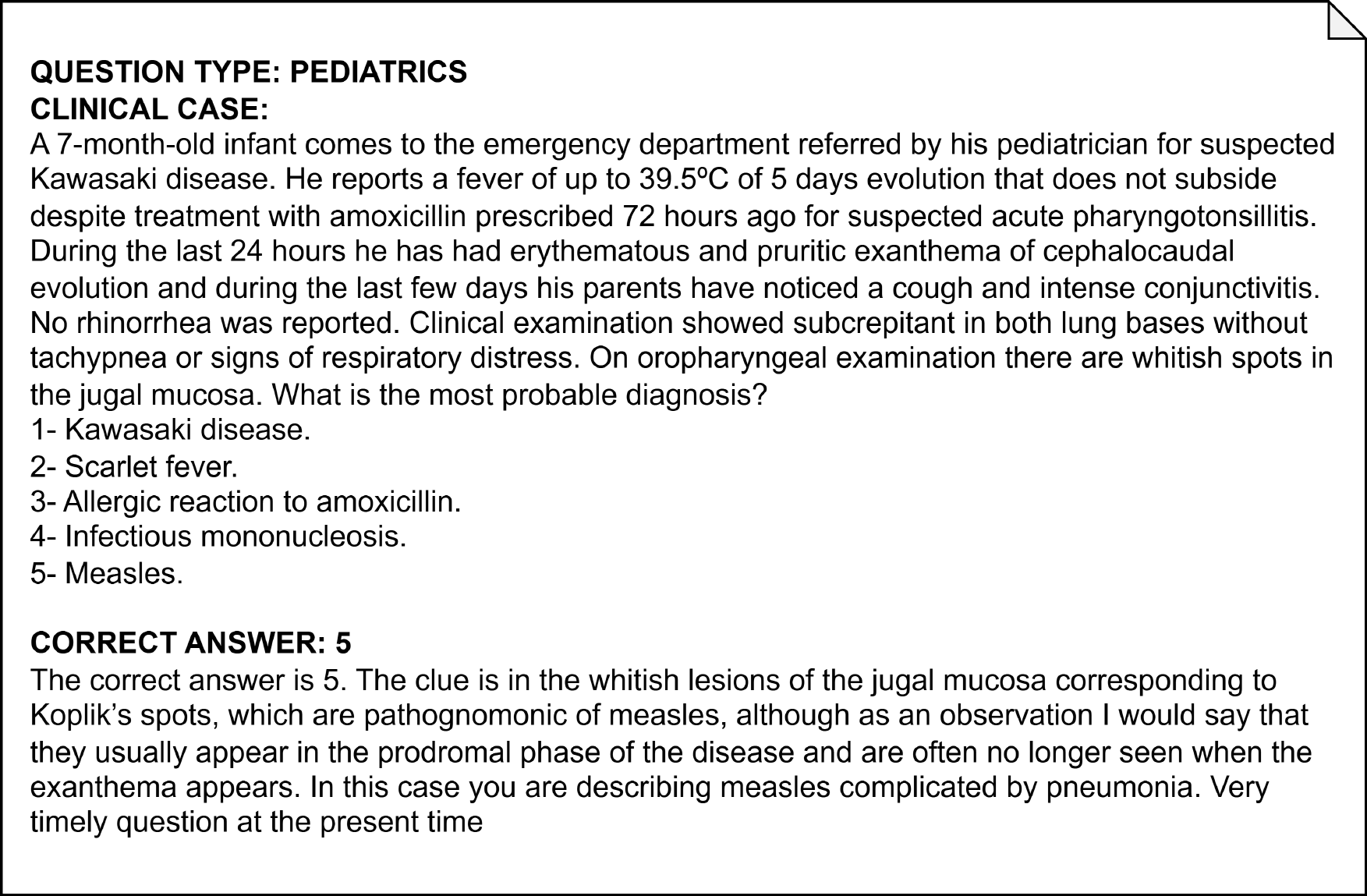}
    \caption{Example of a document structure.}
    \label{fig:clinical case}
\end{figure}

For this task, we have selected the English version of the casiMedicos dataset \cite{sviridova-etal-2024-casimedicos}, specifically designed for argumentative entity recognition using IOB2 annotation, with a subsequently added argument-relation layer. This dataset is particularly well-suited for AM as it is derived from the MIR medical exam, containing clinical cases with an associated question, potential answers (diagnoses or treatments), and explanations justifying the correct response (Figure \ref{fig:clinical case}). This inherent argumentative structure makes it ideal for our study. Out of 953 annotated exam questions, we utilize 553, as these are fully annotated for both steps of AM systems. A total of 2k \textit{premise} and 3.5k \textit{claim} entities, as well as, 760 \textit{attack}, and 825 \textit{support} relations were used in the experiments, divided in train (70\%), dev (10\%) and test (20\%) sets.

AM comprises two steps, each necessitating distinct annotations. For entity recognition (Section \ref{sec:entrec}), we leverage the IOB2 annotations to identify \textit{premises} and \textit{claims} within clinical cases. For entity-relation extraction (Section \ref{sec:relext}), we adopt a CoNLL-format relation annotation that defines argumentative links as either \textit{support} or \textit{attack}. However, in this second step, we consider only relations where an entity directly connects to a potential answer of the exam (hereafter referred to as \textit{Major Claim}), treating the remaining claims of the text as supporting entities that build towards them. This ensures the extraction of arguments that either support or oppose healthcare strategies. This allows us to focus on relations of the form \textit{entity \{X\} \textbf{support/attack} Major Claim \{Y\}}.

\subsubsection{Models} \label{sec:models}

In the first phase, we fine-tune a domain-specific bilingual medical language model \cite{delaiglesia2023eribertabilingualpretrainedlanguage} to identify argumentative entities within clinical cases. These entities, despite being argumentative, are still domain-specific. Thus, using a model with prior medical knowledge ensures more precise identification of technical arguments that might be overlooked by general-purpose models.

For the second step, we utilize a model pre-trained for the NLI task \cite{laurer_less_2022}, leveraging its prior understanding of argumentation for medical-entity relation extraction. Although not explicitly trained in the medical domain, its pre-existing argumentative knowledge is highly valuable and useful for this task. Additionally, we employ a RoBERTa base model \cite{robertabase} to compare its standard fine-tuned text classification performance against the results achieved through our NLI-based approach.

\subsubsection{Ask2Transformers Library}  
The goal of the second step is to determine the relationship between a medical argumentative entity and a potential healthcare strategy. Standard NLI models do not directly classify relationships as \textit{support} or \textit{attack} between two entities; rather, they assess whether a hypothesis (e.g., \textit{X \textbf{support/attack} Y}) holds an entailment, neutral, or contradiction relation given a premise (the clinical case). To address this, we employ Ask2Transformers \cite{sainz2021ask2transformers}, which enables zero-shot domain labelling by performing multiple zero-shot trials and predicting the most probable relationship (\textit{support}, \textit{attack}, or \textit{no-relation}) using various verbalizations to represent each possible relation. To determine the most likely relationship, the verbalization with the highest entailment score is selected. If its probability exceeds a specified threshold, it is classified as the corresponding relation. If it does not surpass the threshold, the output is classified as \textit{no-relation}. On the top of that, once the zero-shot process is completed for all instances, the library provides an optimal threshold for the task.

\subsection{Methods}

\subsubsection{Entity recognition} \label{sec:entrec}
As mentioned in Section \ref{sec:dataset}, the exam answers (\textit{Major Claims}) are treated as goal entities in the second step. However, at this stage, only \textit{premise} and \textit{claim} entities are considered, while the previously mentioned \textit{Major Claim} entity is set aside. These entities remain classified as \textit{claims} in this step, as it aligns with their true categorization, even if afterwards they will be treated separately, since the overarching goal of this AM system is to identify arguments supporting or opposing a healthcare strategies (e.g., diagnosis, treatment, or procedure). As previously noted, these entities will be extracted using a token classification method.

\subsubsection{Entity-relation extraction}\label{sec:relext}

The first step in this phase involves fine-tuning an NLI model with prior knowledge of general argumentation, for application in the medical domain. To achieve this, we transform the existing annotations, which specify argumentative relations between entities, into a format compatible with NLI structure. 

\begin{figure}[ht!]
  \centering
  \includegraphics[width=0.9\linewidth]{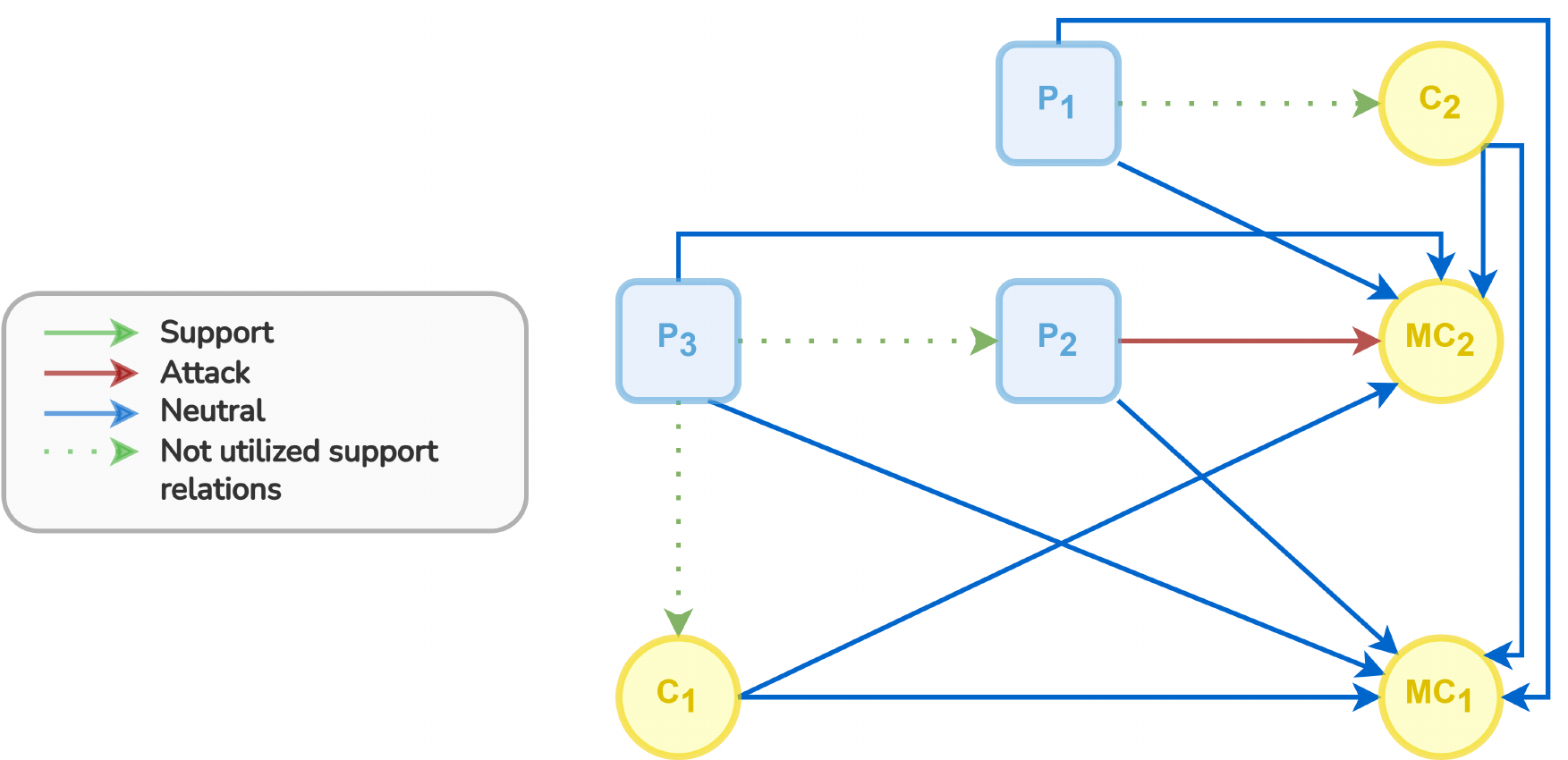}
  \caption{Graph representing \textbf{V1}, where any entity-to-\textit{Major Claim} relation not already classified is considered neutral. Unused annotated relations are represented with a dot arrow, as in Figure \ref{fig:grafo}, Figure \ref{fig:erpinak} and Figure \ref{fig:sueltas}.}
  \label{fig:todasneutras}
\end{figure}

Specifically, we remove potential answers from the exam, using the remaining clinical case and explanation as the premise. The hypothesis is then formulated using argument tuples in the format \textit{X \textbf{support/attack} Y}, where Y always represents a \textit{Major Claim}. Moreover, we remove in this process some annotated entities, such as ``The correct answer is 1", which are considered argumentative entities, but do not provide any relevant insight even if they are directly connected to the \textit{Major Claims}. To assign classifications to these premise-hypothesis pairs, we apply the following criteria: existing annotated relations will be treated as \textit{entailments}, and \textit{contradictions} the opposite of the annotated relations. For the \textit{neutrals}' case, it is not clear which format suits best the neutrality between the entities, thus we follow four distinct approaches to determine the best neutrality representation for our data. The types are:

\paragraph{1st Variation (V1)} Any entity-to-\textit{Major Claim} relation not already classified is considered neutral (Figure \ref{fig:todasneutras}). 

\begin{figure}[ht!]
  \centering
  \includegraphics[width=0.9\linewidth]{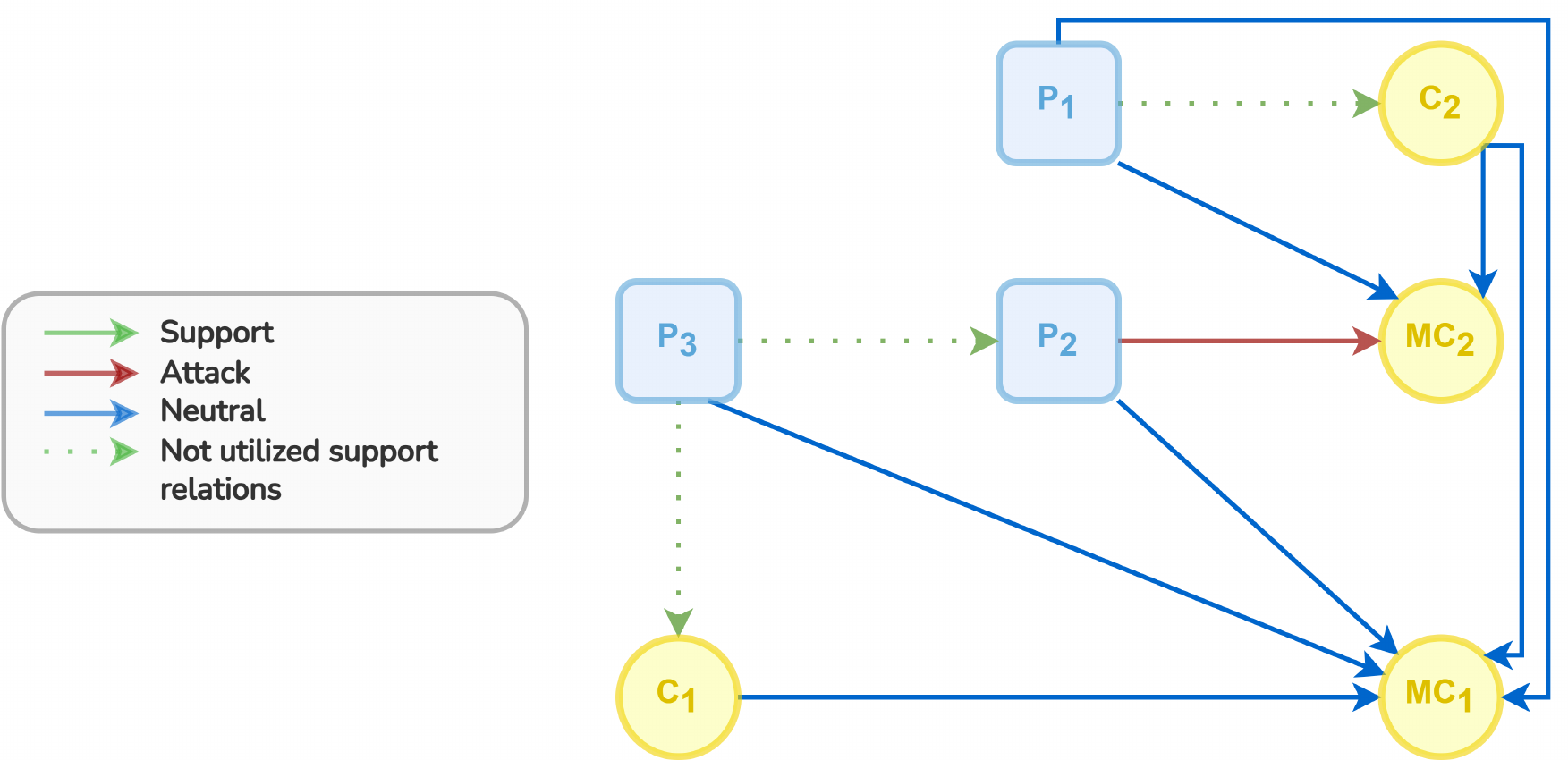}
  \caption{Illustration of a \textbf{V2} graph. Two entities cannot be within the same sub-graph, to be considered neutral. The three sub-graphs in the image: $P_1$-$C_2$; $P_2$-$P_3$-$C_1$-$MC_2$ and $MC_1$ alone.}
  \label{fig:grafo}
\end{figure}

\paragraph{2th Variation (V2)} Only relations between completely unconnected entities (considering relation transitivity) within the relation graph are neutral (Figure \ref{fig:grafo}).

\begin{figure}[ht!]
  \centering
  \includegraphics[width=0.9\linewidth]{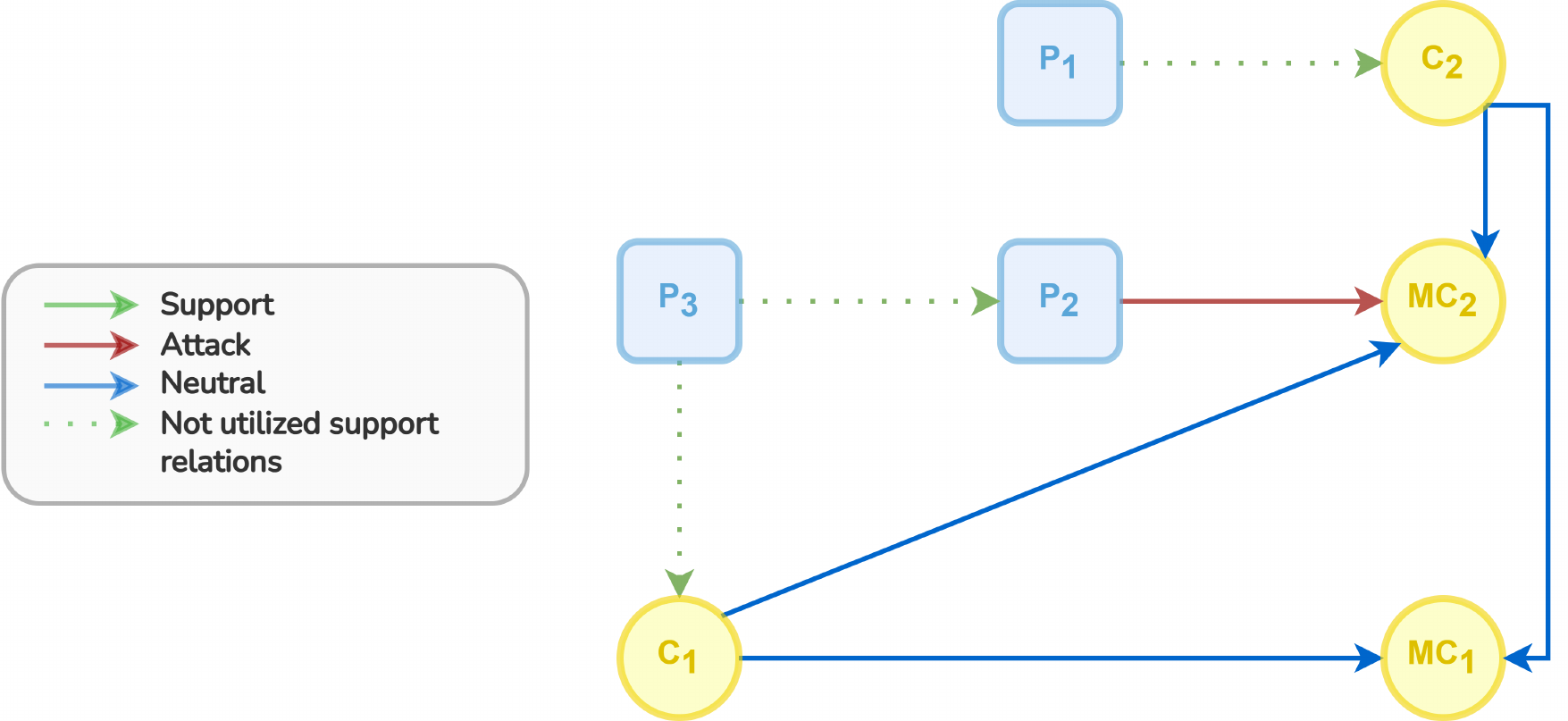}
  \caption{ Representation of \textbf{V3}, where a relation is neutral only if X (\textit{X \textbf{support/attack} Y}) does not appear as X in any other relation. For instance, $C1$ is not the X of any other relation.}
  \label{fig:erpinak}
\end{figure}

\paragraph{3th Variation (V3)} A relation is neutral only if X (\textit{X \textbf{support/attack} Y}) does not appear as X in any other relation (Figure \ref{fig:erpinak}). 

\begin{figure}[ht!]
  \centering
  \includegraphics[width=0.9\linewidth]{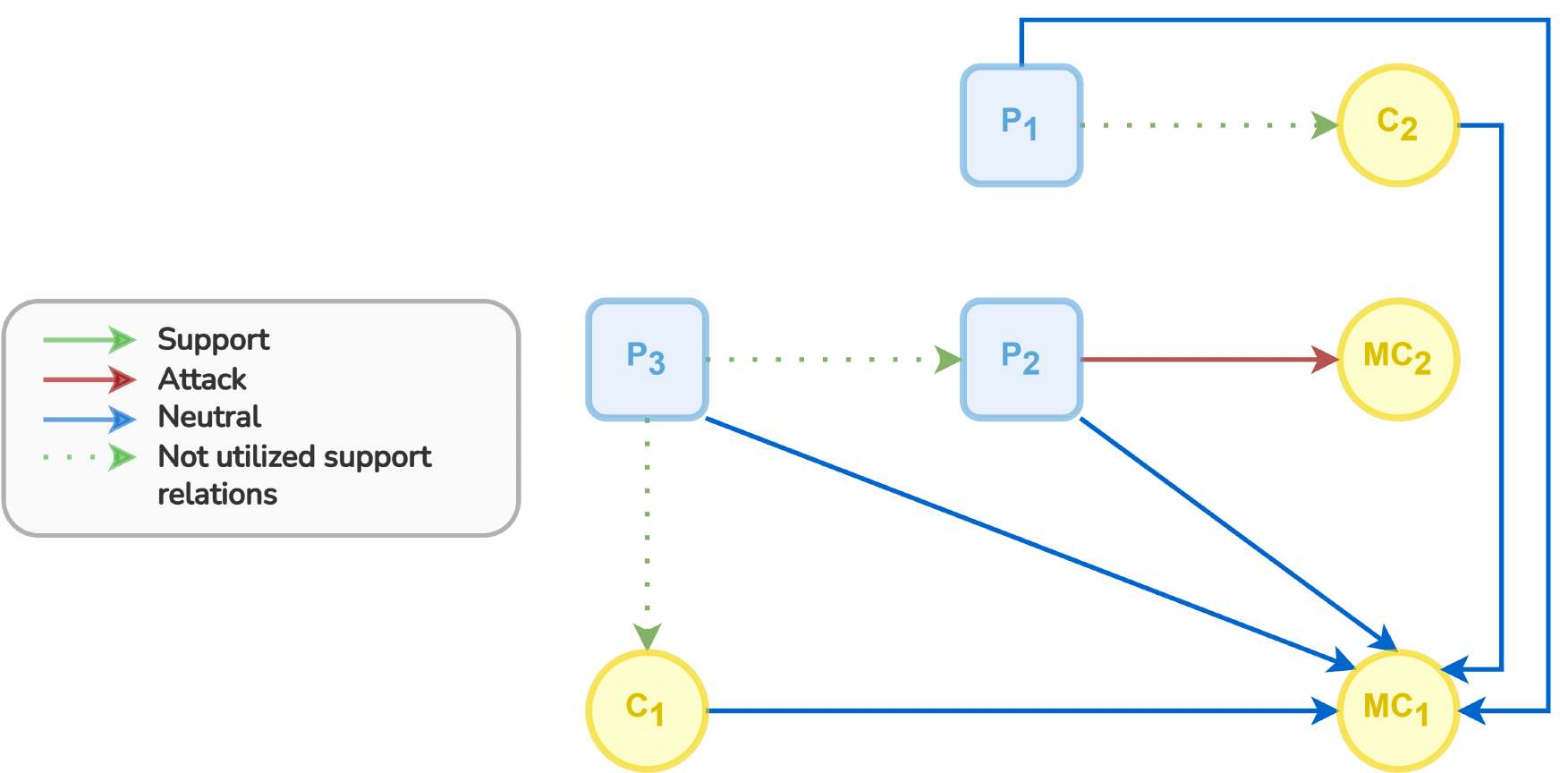}
  \caption{Drawing of \textbf{V4}, in which at least one entity must have no other relation with any other entity to be considered part of a neutral relation. Only $MC1$ in this case.}
  \label{fig:sueltas}
\end{figure}

\paragraph{4th Variation (V4)} At least one entity in the relation must have no other relation with any other entity in the whole relation graph (Figure \ref{fig:sueltas}). 

Since neutral relation cases significantly outnumber the other two, their occurrences in the training set are balanced to ensure an equal distribution of \textit{entailment}, \textit{contradiction}, and \textit{neutral} instances. \textit{Entailment} and \textit{contradiction} always appear in equal amounts, as \textit{contradictions} are generated by inverting \textit{entailments}. To promote greater diversity in the premises (clinical cases) of neutral instances during model fine-tuning, a uniform number of neutral cases is selected from each clinical text, up to the predefined maximum limit. However, the instances' label imbalance will be maintained in both development and test sets.

In addition to refining these four data configurations, we experiment with different verbalizations for the possible relations. For example, an \textit{attack} relation might be better represented with terms like ``refute" or ``contradict" rather than simply ``attack''. Consequently, for a reference point we use the ``support" and ``attack" verbalizations themselves and then we explore the following alternative verbalizations: ``attack", ``challenge", ``contradict", ``dispute", ``refute" for \textit{attack}, and ``support", ``confirm", ``corroborate", ``endorse", ``validate" for \textit{support}.

\begin{figure}[ht]
    \centering
    \includegraphics[width=0.85\linewidth]{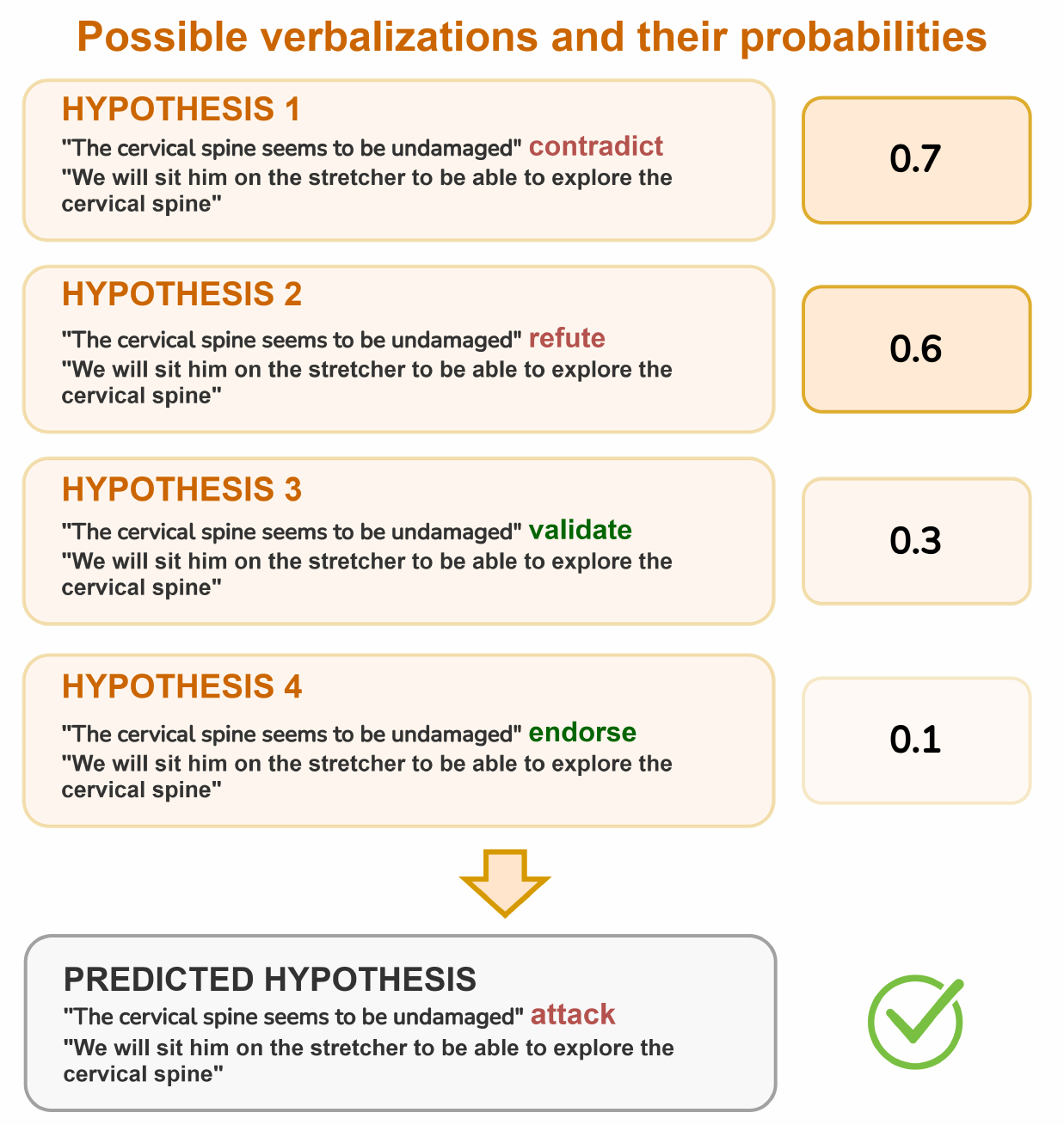}
    \caption{Given two sentences to the Ask2Transformers system---namely, ``The cervical spine seems to be undamaged" and ``We will sit him on the stretcher to be able to explore the cervical spine"---, it performs a zero-shot against the fine-tuned NLI model, storing the entailment probability for each verbalization. Then selects the one with the highest entailment probability and assigns the corresponding relation-label to the input sentences.}
    \label{fig:verbalizationselec}
\end{figure}

Following fine-tuning with these verbalizations, we evaluate their effectiveness by analysing the results produced by Ask2Transformers. As illustrated in Figure \ref{fig:verbalizationselec}, for each pair of arguments, the library tests all possible verbalization combinations, selecting the one with the highest entailment probability before assigning a corresponding label. This approach allows us to identify the most effective verbalizations for accurate classification, which will be retained for the final model implementation.

Through these two-step experiments for relation extraction, we aim to demonstrate that leveraging the prior argumentative knowledge of NLI models can yield superior results compared to end-to-end text classification methods, particularly in data-scarce environments. To substantiate this, as mentioned in Section \ref{sec:models}, we compare the results of our two-step approach with its one-step text classification counterpart across different data availability scenarios, specifically using 5\%, 15\%, 20\%, and 100\% of the dataset.

\section{Results}

In this section, we present the experimental results obtained from the fine-tuned token classification model for entity recognition and NLI-based entity-relation extraction system.
\subsection{Entity recognition}

For the entity recognition model, hyperparameter tuning explored various learning rates (1e-5, 25e-6, 5e-5, 75e-6) and weight decays (0.0, 0.001, 0.01), with a constant warm-up rate of 0.05, 15 epochs, and batch size of 32. The best performance was achieved at a learning rate of 5e-5 and a weight decay of 0.01, yielding a Micro-F1 score of 0.8292, with Premise-F1 of 0.8307 and Claim-F1 of 0.8285. This slightly outperformed the 75e-6 learning rate model, which achieved a Micro-F1 of 0.8275, also with a weight decay of 0.01. 

\begin{table}[ht]
\centering
\begin{tabular}{cccc}
\hline\rule{-2pt}{15pt}
LRate & P-F1 & C-F1 & Micro-F1\\ \hline\rule{-2pt}{10pt}
1e-5                   & 0.7305              & 0.7982            & 0.7503            \\
25e-6                  & 0.7850              & 0.8000            & 0.7956            \\
5e-5                   & \textbf{0.8307}     & \textbf{0.8285}   & \textbf{0.8292}   \\
75e-6                  & 0.8298              & 0.8266            & 0.8275            \\ \hline
\end{tabular}
\caption{\label{ner_results}Performance of the model for different learning rates and weight decay of 0.01. The table reports F1 scores for \textit{premises} (P-F1), \textit{claims} (C-F1), and overall Micro-F1 on development set. The best results are highlighted in bold.}
\end{table}

Across most parameter settings, Claim-F1 was higher than Premise-F1, as expected since the dataset has roughly twice as many \textit{claim} entities as \textit{premise} entities. With higher learning rates leading to a more balanced performance. Table \ref{ner_results} details the hyperparameter results.

To evaluate model stability, multiple runs were conducted with the optimal hyperparameters and different random seeds. The results showed consistent performance, with Claim-F1 outperforming Premise-F1. However, on the test set, performance slightly decreased, with Micro-F1 dropping from 0.8292 to 0.7978. Premise-F1 fell to 0.7670, while Claim-F1 to 0.8020 (Table \ref{ner_test_results}).

\begin{table}[ht]
\centering
\begin{tabular}{cccc}
\hline\rule{-2pt}{15pt}
Set  & P-F1 & C-F1 & Micro-F1 \\ \hline\rule{-2pt}{15pt}
Dev  & 0.8307     & 0.8285   & 0.8292   \\ 
Test & 0.7670     & 0.8020   & 0.7978   \\ \hline
\end{tabular}
\caption{\label{ner_test_results}Performance of the optimal entity recognition model on development and test.}
\end{table}

 To investigate the impact of text structure on model performance, additional experiments were performed by training separate models on distinct sections of the texts. As presented in Table \ref{tab:ner_div_results}, the model trained exclusively on question-answer sections achieved a Micro-F1 of 0.9755, while the model trained only on final explanation achieved a Micro-F1 of 0.6478.

\begin{table}[ht]
\centering
\begin{tabular}{cccc}
\hline\rule{-2pt}{15pt}
Section   & P-F1 & C-F1 & Micro-F1 \\ \hline\rule{-2pt}{15pt}
Q\&A   & 0.9457              & 0.9980    & 0.9755                 \\ 
Explanation & 0.3704              & 0.6858    &0.6478               \\ \hline
\end{tabular}
\caption{Comparison of entity recognition performance on development set when training separate models on different sections of clinical texts: question and possible answer sections (Q\&A) versus explanation sections.}
\label{tab:ner_div_results}
\end{table}

\subsection{Entity-relation extraction}

As outlined in Section \ref{sec:relext}, Ask2Transformers uses the entailment probability of each verbalization to select the most likely one and classify the relationship between the two argumentative entities. Thus, fine-tuning of the NLI models in this step is based on the entailment F1-score. In the hyperparameter tuning we tested two learning rates (5e-5, 75e-6), weight decays (0.001, 0.01), warm-up rates (0.0 and 0.05), varying epochs (2 to 10), and a constant batch size of 16.

With the aim of clarifying the process focusing only on the impact different neutral selection strategies have, we start by using a simplified version of the NLI process where we employ a single verbalization for \textit{support} and a single verbalization of \textit{attack}. As shown in Table \ref{table:baseline_res}, the best performance on the development set was achieved by the V4 dataset, with a learning rate of 75e-6, followed closely by the V3 dataset at 5e-5, showing almost a 0.08-point gap compared to the other two datasets. All top results were obtained with a weight decay of 0.01 and a warm-up rate of 0.05.
\begin{table}[ht]
\centering
\begin{tabular}{cccc}
\hline\rule{-2pt}{15pt}
Dataset & LR& F1-Entailment & $\sigma^2$ \\ \hline\rule{-2pt}{10pt}
V1    &5e-5& 0.1819        & ±0.0164 \\
V2    &5e-5& 0.1879        & ±0.0086 \\
V3    &5e-5& 0.2650        & ±0.0248 \\
V4    &75e-6& \textbf{0.2653} & ±0.0205 \\ \hline
\end{tabular}
\caption{\label{table:baseline_res}Final entailment F1-score results of the simplified version of the NLI model on development data---using only ``support" and ``attack" verbalizations. All the results were obtained with a 0.01 weight decay and 0.05 of warm-up. The best result is highlighted in bold.}
\end{table}

\begin{figure*}[ht]
    \centering
    \includegraphics[width=0.8\linewidth]{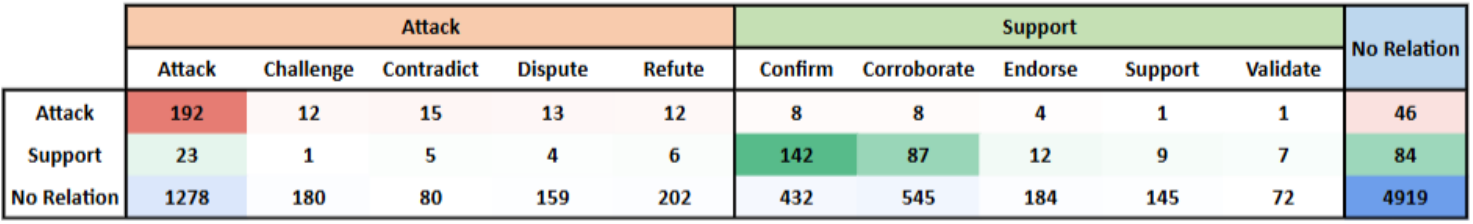}
    \caption{Usage of the tested verbalizations to predict argumentative-entity relation. The verbalizations are ``attack", ``challenge", ``contradict", ``dispute", ``refute" for \textit{attack} relation, and ``support", ``confirm", ``corroborate", ``endorse", ``validate" for \textit{support}.}
    \label{fig:verbalization_matrix}
\end{figure*}
After fine-tuning and evaluating the models, their performance on the classification task was assessed using Ask2Transformers' zero-shot approach. Initially, a threshold of 0.2 was set to classify a \textit{no-relation} outcome when no verbalizations exceeded it. We performed a zero-shot classification on the development set, followed by the application of the optimal threshold on the test set for final results. As shown in Table \ref{table:baseline_res_inf}, the V1 model achieved the best performance (\textit{attack} and \textit{support} mean F1-score) on both sets, despite having the lowest scores in the reference point experiments (Table \ref{table:baseline_res}). However, it required a significantly higher threshold to classify a relation compared to the other models, particularly V4, which needed only a 0.3523 threshold.

\begin{table}[ht]
\centering
\begin{tabular}{cccc}
\hline\rule{-2pt}{15pt}
Dataset & F1-Dev & Threshold & F1-Test \\ \hline\rule{-2pt}{10pt}
V1    & \textbf{0.3628}    &  0.9830   &   \textbf{0.3255}  \\
V2    & 0.3039    &  0.9949   &   0.2944 \\
V3    & 0.3279    &  0.7507   &   0.3236 \\
V4    & 0.3273    &  0.3523   &   0.3039 \\ \hline
\end{tabular}
\caption{\label{table:baseline_res_inf}Final F1-score results for \textit{attack} and \textit{support} relations in the inference process on development and test data, using the model trained on initial verbalizations. Best results highlighted in bold.}
\end{table}

Next, we tested other verbalizations mentioned in Section \ref{sec:relext} to identify the most effective ones. To do so, we tested the newly trained models on the development set and identified which verbalization was used for each prediction, as shown in the confusion matrix in Figure \ref{fig:verbalization_matrix}. We concluded that the most effective verbalization for the \textit{attack} relation was ``attack'' itself, while for \textit{support}, both ``confirm" and ``corroborate" verbalizations proved to be crucial for accurate predictions. As a result, these verbalizations were selected for the final model.

When training with the selected verbalizations and following the same evaluation process, the results in Table \ref{table:final_res_inf} show that the V4-based model remained the top performer. However, it failed to outperform the initial version of V1 model, despite demonstrating greater stability, maintaining consistent performance across verbalization variations. In contrast, the V1-based model experienced a performance drop, with a 0.05-point decrease in the test set and a 0.10-point drop in the development set.

\begin{table}[ht]
\centering
\begin{tabular}{cccc}
\hline\rule{-2pt}{15pt}
Dataset & F1-Dev & Threshold & F1-Test \\ \hline\rule{-2pt}{10pt}
V1    & 0.2659    &  0.9519   &   0.2789  \\
V2    & 0.3029    &  0.9949   &   0.2903 \\
V3    & 0.3200    &  0.8839   &   0.3145 \\
V4    & \textbf{0.3252}    &  0.8579   &   \textbf{0.3205} \\ \hline
\end{tabular}
\caption{\label{table:final_res_inf}Final F1-score results for \textit{attack} and \textit{support} relations in the inference process on development and test data, using the model trained on final verbalizations. Best results highlighted in bold.}
\end{table}

\begin{figure*}[ht]
    \centering
    \includegraphics[width=0.87\linewidth]{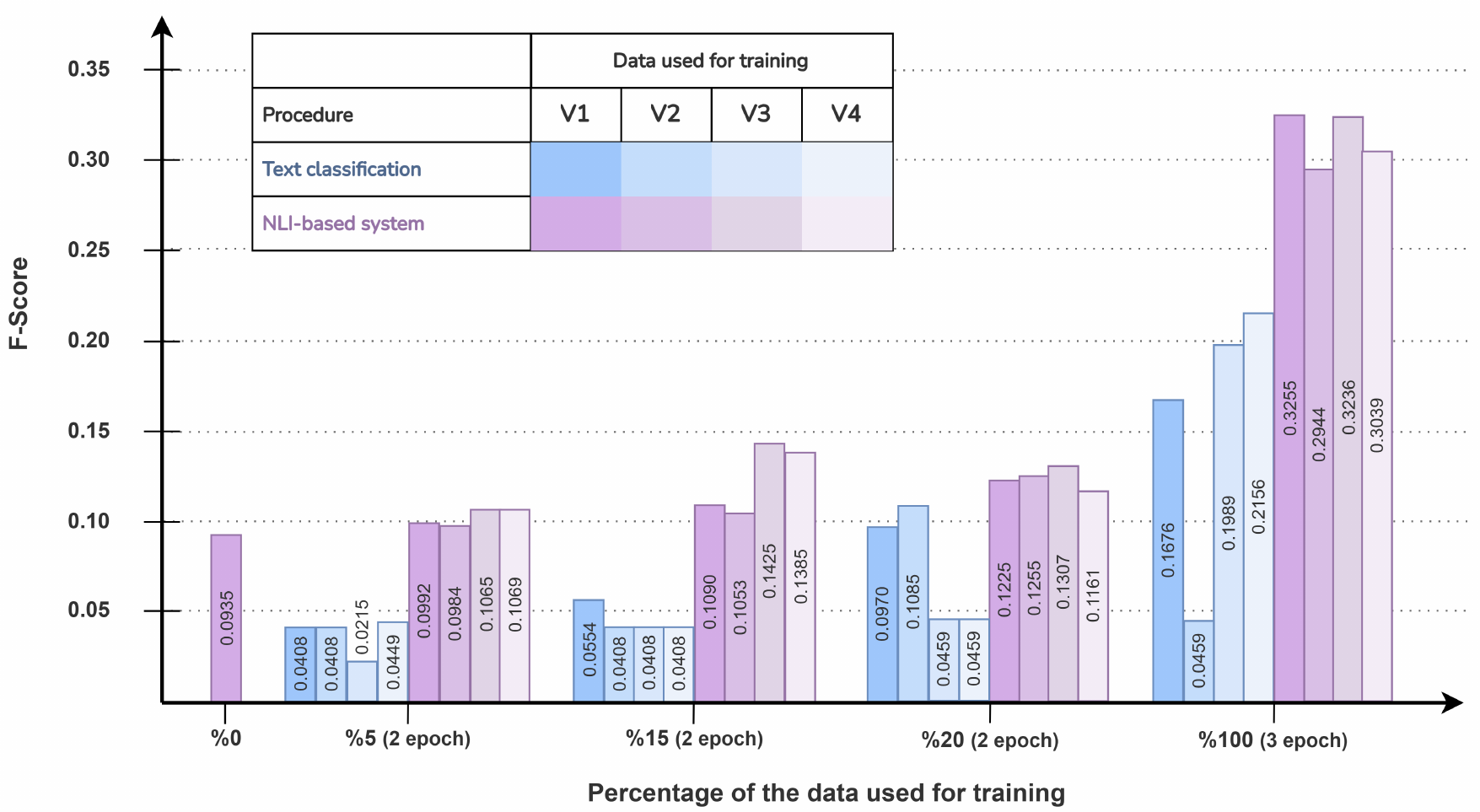}
    \caption{Results on the test dataset for predicting the most probable relation---\textit{attack}, \textit{support}, or \textit{no-relation}---between medical argumentative-entity pairs. Initial verbalizations are employed and the plotted metric represents the mean F1-score for the \textit{attack} and \textit{support} relations. Less epochs are used in smaller sets to avoid overfitting.}
    \label{fig:ENG_grafico_final}
\end{figure*}

After establishing initial verbalizations as the best performing ones, we conducted experiments in data-scarce situations, comparing the NLI-based models with their text classification counterparts. These experiments used 5\%, 15\%, 20\%, and 100\% of the training data, with a zero-shot evaluation using the NLI model before fine-tuning to assess its pre-fine-tuning performance in medical domain argumentation (Figure \ref{fig:ENG_grafico_final}). In the cases with less data, we used 2 epochs instead of 3 to avoid overfitting.

\section{Discussion}

Building upon the experimental results, this section analyses the effectiveness and limitations of the proposed approach to argumentative entity recognition and entity-relation extraction in clinical texts.
\subsection{Entity recognition results analysis} \label{sec:entrec_erran} 
 
The results demonstrate the viability of transformer-based token classification for identifying argumentative entities in clinical texts, achieving a Micro-F1 of 0.7978 on the test set. However, the performance gap between structured question-answer sections and less structured explanatory paragraphs highlights a notable challenge. While the model effectively leveraged structure when present, it struggled to capture argumentative entities in more natural, free-flowing discourse. This observation aligns with previous research emphasizing the crucial role of document structure in argument identification \cite{teufel2009towards}.

Error analysis revealed several factors affecting performance. First, a positional bias was observed, since entities in early paragraphs were identified more accurately than those later on. This likely stems from exam-like sections, where \textit{premises} appear in the statement and \textit{claims} in the answers, unlike later explanations where both entity types appear together. Additionally, the model faced boundary detection issues particularly around conjunctions such as \textit{but} and \textit{and}, often causing entity merging, whereas subordinating conjunctions like \textit{although}, \textit{because}, and \textit{where} frequently led to entity splitting. Finally, some entities were fragmented, with individual words incorrectly classified as stand-alone components rather than parts of larger argumentative structures. These challenges align with findings from human annotation studies \cite{enhancingmayer2021}, highlighting the complexity of precisely demarcating argumentative entities, whether performed by humans or machines. 

Placing this work in a broader context, the study extends Argument Mining (AM) research to clinical texts, addressing challenges unique to medical discourse. While previous studies have focused on the MEDLINE database and PICO entities \cite{stylianou2021transformed,mayer2020transformer}, medical argumentation shows distinct patterns across different clinical texts due to the interplay between diagnostic reasoning and Evidence-Based Medicine. Unlike in MEDLINE, argumentative entities in casiMedicos appear in exam-like contexts, particularly in the explanatory final section. These texts, aimed at students with medical knowledge, use a more academic tone and concise arguments, making \textit{premise} and \textit{claim} identification more challenging. Nonetheless, the results show that transformer-based token classification approaches can effectively capture argumentative structures, even in these complex scenarios.

Future work should focus on enhancing boundary detection, particularly in relation to connective words, and refining the handling of less structured explanatory texts. Additionally, integrating domain-specific knowledge of clinical reasoning patterns could further improve performance in medical Argument Mining \cite{stylianou2021transformed}.

\subsection{Entity-relation extraction results analysis}\label{sec:relext_erran}

The entity-relation extraction results (Section \ref{sec:relext}) reveal that models trained with the simplified version of verbalizations---``support" and ``attack"---generally outperform those with final verbalizations, except for those trained on the V4 dataset. A key difference between V1 and V4 models arises from their contrasting criteria for defining neutral relations. V1 considers all non-annotated relations as neutral, ignoring potential transitive relations, while V4 defines only completely isolated entities as neutral. Interestingly, each approach produces the best results in different contexts---V1 excels with initial verbalizations, while V4 performs better with final verbalizations.

Furthermore, while V1 models show better performance with the initial version, V4 models demonstrate greater stability, with a 0.02 F1-score improvement on test data, compared to a 0.05 point drop in V1 models when using the final verbalizations. This suggests that a more restrictive approach to defining neutrality, like that of V4, leads to more stable results with less variation. This stability is further supported by the performance of V3, which also uses a relatively restrictive neutrality criterion and shows consistent and competitive results. These findings highlight the significant impact the criteria for defining neutral cases can have on model performance, illustrating the complexity involved in entity-relation extraction tasks already observed in previous works like \cite{eger2017neural}. 

Overall results show that using an NLI approach improves performance compared to straightforward text classification models, especially when dealing with limited data. The NLI model consistently outperforms the non-fine-tuned version, with the best improvement reaching 0.232 points. It also surpasses the text classification model by 0.1151 points when using all the data and running for three epochs. Notably, the NLI model is more stable, showing smaller performance fluctuations across different data sizes, while the text classification model's results vary significantly, particularly when using 20\% and 100\% of the data.

These findings suggest that NLI is highly beneficial in fields like medicine, where data is often scarce, as it enables models to leverage the previous knowledge in general argumentation to extract more value from limited domain-specific data. Additionally, with more data and continued experimentation, NLI based systems show great potential for accurately identifying and distinguishing argumentative relationships in clinical texts. This stability and effectiveness make NLI a promising approach for improving performance in complex language tasks, particularly in data-constrained environments.

\section{Conclusions}

This work presents a medical-domain AM system trained on the casiMedicos dataset, demonstrating the effectiveness of token classification methods for argumentative entity recognition \textbf{(RQ1)}, as well as the benefits of using intermediate tasks like NLI to enable cross-task transfer learning and improve entity-relation identification in low-resource, specialized domains such as medicine \textbf{(RQ2)}. Thanks to this cross-task transfer learning, the proposed methodology outperforms a basic text classification baseline \textbf{(RQ3)}, especially in low-data scenarios.

Future approaches using this techniques may include for entity recognition, the focus on enhancing boundary detection---which is also difficult to achieve in human annotation---, as well as refining the handling of less structured texts, integrating domain-specific knowledge of clinical reasoning patterns. As for relation identification, leveraging other clinical texts with different structures and arguments would also be beneficial in this step to measure the performance of this methodology across different texts. It would also be necessary to integrate both parts of the pipeline---entity recognition and relation identification---since they were tested separately in this study. This integration will enable a more comprehensive evaluation of the full system's performance and its applicability in real-world clinical scenarios.

In conclusion, our work lays the groundwork for future tools that provide evidence-based justifications for machine-generated clinical conclusions leveraging Argument Mining together with previous general argumentative knowledge to improve domain-specific results.

\section*{Acknowledgements}
This work has been partially supported by the HiTZ Center and the Basque Government (IXA excellence research group funding IT-1570-22 and IKER-GAITU project), as well as by the Spanish Ministry of Universities, Science and Innovation MCIN/AEI/10.13039/501100011033 by means of the projects: Proyectos de Generación de Conocimiento 2022 (EDHER-MED/EDHIA PID2022-136522OB-C22), DeepKnowledge (PID2021-127777OB-C21), DeepMinor (CNS2023-144375), and EU NextGeneration EU/PRTR (DeepR3 TED2021-130295B-C31). And also by an FPU grant (Formación de Profesorado Universitario) from the Spanish Ministry of Science, Innovation and Universities (MCIU) to the third author (FPU23/03347).

\bibliographystyle{fullname}
\bibliography{paper}

\end{document}